\documentclass{article}

\usepackage[preprint]{corl_2026} 

\usepackage{microtype}
\usepackage{graphicx}
\usepackage[export]{adjustbox}
\usepackage{subcaption}
\usepackage{booktabs}
\usepackage{amsmath}
\usepackage{amssymb}
\usepackage{mathtools}
\usepackage{amsthm}

\usepackage{array}
\usepackage{multirow}
\usepackage{pifont}
\usepackage{marvosym}
\usepackage[table]{xcolor}
\usepackage{pgfplots}
\usepackage{float}
\usepackage{placeins}
\usepackage[capitalize,noabbrev]{cleveref}
\usepackage{caption}
\usepackage{wrapfig}
\usepackage{algorithm}
\usepackage{algpseudocode}
\algrenewcommand\algorithmiccomment[1]{\hfill$\triangleright$~#1}

\pgfplotsset{compat=1.18}

\newcommand{\cmark}{\ding{51}}
\newcommand{\xmark}{\ding{55}}
\newcommand{\leadmark}{\textsuperscript{\dag}}
\newcommand{\corrmark}{\textsuperscript{\ding{41}}}

\theoremstyle{plain}

\theoremstyle{definition}

\theoremstyle{remark}

\title{MotionVLA: Injecting Geometric Motion into Vision-Language-Action Model}

\hypersetup{
  pdftitle={MotionVLA: Injecting Geometric Motion into Vision-Language-Action Model},
  pdfauthor={Shanglin Yuan, Weiheng Zhao, Xianda Guo, Wei Sui, Li Yu, Wenyu Liu, Xinggang Wang},
  pdfkeywords={Vision-Language-Action, Motion History, Robot Learning}
}

\author{
  \textbf{Shanglin Yuan\textsuperscript{1,2} \quad
  Weiheng Zhao\textsuperscript{1,2} \quad
  Xianda Guo\textsuperscript{2,3}}\\[-0.2mm]
  \textbf{Wei Sui\textsuperscript{2}\leadmark \quad
  Li Yu\textsuperscript{1} \quad
  Wenyu Liu\textsuperscript{1} \quad
  Xinggang Wang\textsuperscript{1}\corrmark}\\[1mm]
  \textsuperscript{1}Huazhong University of Science and Technology \quad
  \textsuperscript{2}D-Robotics\\[-0.2mm]
  \textsuperscript{3}Wuhan University
}

\usepackage{hyperref}

\hypersetup{
  pdftitle={MotionVLA: Injecting Geometric Motion into Vision-Language-Action Model},
  pdfauthor={Anonymous Author(s)},
  pdfsubject={},
  pdfkeywords={Vision-Language-Action, Motion History, Robot Learning},
  pdfcreator={},
  pdfproducer={}
}

\begin{document}
\maketitle


\begin{abstract}
Vision-language-action (VLA) models increasingly condition robot policies on 
history, depth, or 4D features to resolve ambiguity in long-horizon manipulation. 
However, more spatiotemporal evidence is not necessarily better: when the injected 
evidence is not motion-consistent, it can introduce geometric drift, fragmented 
temporal cues, and unstable action generation. This raises a simple question: 
should a VLA remember past frames, or remember the motion that connects them? 
We introduce \textbf{MotionVLA}, a motion-history interface that converts a short 
past-only video window into compact, time-continuous trajectory-field tokens. 
Instead of treating history as a sparse set of independently lifted frames, 
MotionVLA represents recent observations as physically coherent motion evidence. Current visual tokens query this history to retrieve task-relevant motion information, which is then recoupled into the VLA stream under trajectory-grounded supervision. 
Experiments across simulation benchmarks and preliminary real-robot rollouts show 
that MotionVLA improves long-horizon manipulation while producing smoother and 
more direct executions. These results suggest that effective VLA memory is not 
just about providing more 4D context, but about exposing motion-consistent evidence 
that is usable for control.
\end{abstract}


\section{Introduction}

Long-horizon manipulation requires a robot policy to infer not only what is visible
now, but also how the robot arrived there. In many multi-stage tasks, the same
current observation can correspond to different control states depending on recent
motion: a robot may need to continue the current subgoal, terminate it, or move on
to the next one. Such visual aliasing makes purely reactive policies fragile and
may trigger the state chaos phenomenon~\cite{MemoryVLA,4DVLA,SwiftVLA}. 
This suggests that memory is not merely additional context for a robot policy, but
a control interface whose structure determines whether past evidence can be used
stably for action generation.

Vision-language-action (VLA) models~\cite{RDT1B,OpenVLA,pi0,pi0FAST,CogACT,SmolVLA,SpatialVLA,CoTVLA,TraceVLA,pi05,OpenVLAOFT,DepthVLA,GeoVLA,SwiftVLA}
provide a strong framework for language-conditioned robot control, formulating
action generation as conditional sequence prediction from visual observations,
proprioception, and natural-language instructions. A common design reuses a
pretrained vision-language model for perception and instruction following, and
attaches an action head for low-level continuous control. For example,
$\pi_0$~\cite{pi0} builds on PaliGemma-3B~\cite{PaliGemma}, inheriting semantic
priors from large-scale vision-language pretraining while adapting them to robot
actions. To reduce ambiguity in long-horizon manipulation, recent methods
increasingly condition VLA policies on additional temporal or spatial evidence,
including history or memory modules~\cite{MemoryVLA,RT1,CronusVLA,TraceVLA,MindExplore},
explicit 3D geometry~\cite{DepthVLA,GeoVLA}, and spatiotemporal extensions that
expose 4D evidence to the policy~\cite{4DVLA,SwiftVLA}.

\begin{figure}[t]
  \vspace{-3mm}
  \centering
  \includegraphics[max width=1\linewidth, trim=0cm 0cm 0cm 0cm, clip]{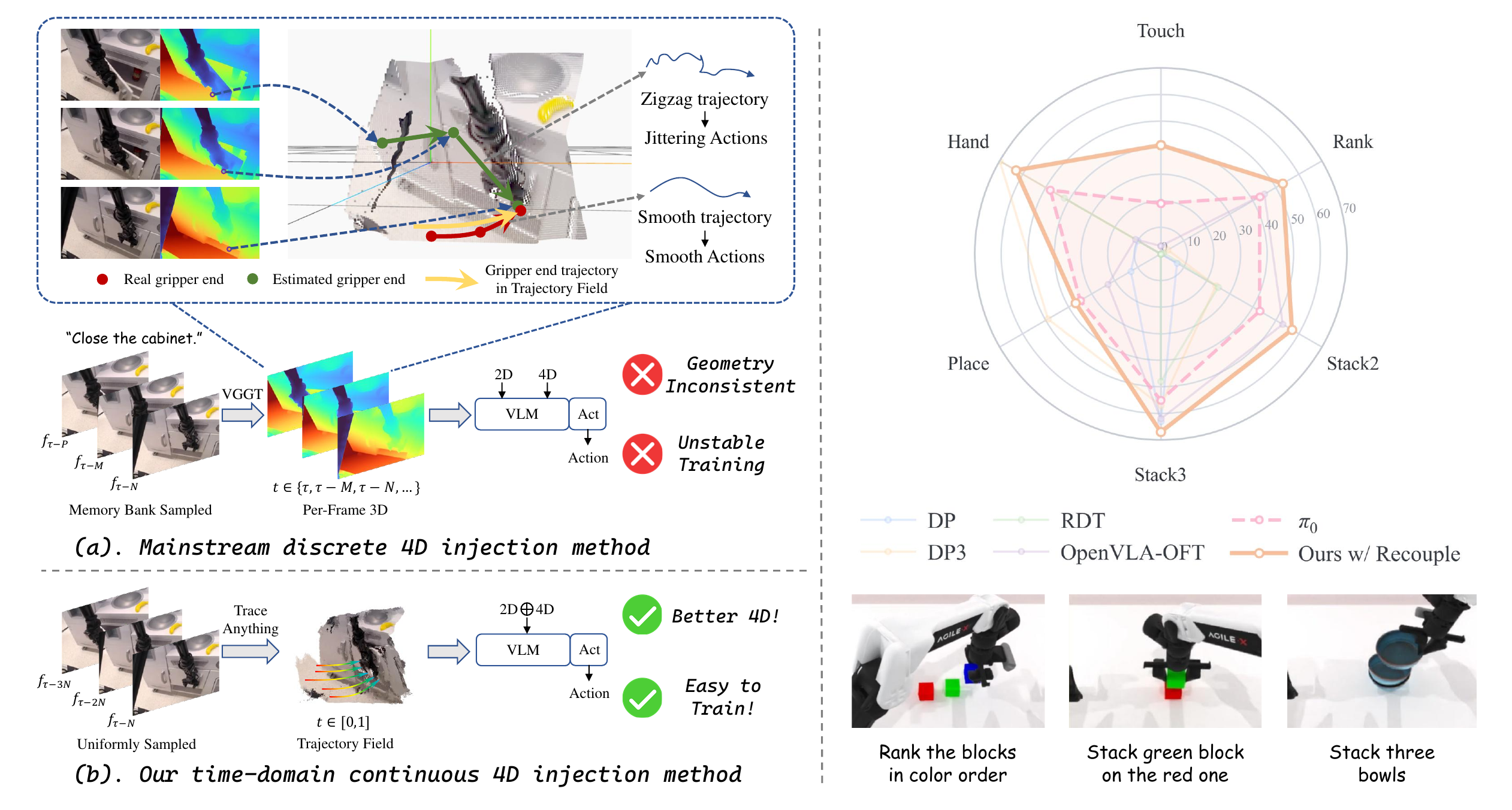}
  \vspace{-2mm}
  \caption{Discrete 4D evidence can be fragmented, while trajectory fields provide a more consistent motion interface. Instead of treating history as a sparse set of independently lifted frames, MotionVLA represents recent observations as queryable, time-continuous motion evidence for smoother and more direct control.}
  \label{fig:teaser} 
  \vspace{-6mm}
\end{figure}

However, the trend of injecting more spatiotemporal evidence hides an important
assumption: the injected evidence must itself be coherent enough to be useful for
control. Our analysis suggests that this assumption does not always hold. In
representative 4D injection pipelines, a policy may still complete an episode
while taking noticeably less direct end-effector paths. As shown in
\autoref{tab:path_efficiency}, this behavior appears in our LIBERO
path-efficiency analysis, where the executed trajectories of a 4D-injected VLA
variant can contain substantial detours relative to expert demonstrations and the
$\pi_0$ baseline. This observation suggests that the bottleneck is not simply
whether a VLA has access to history or geometry, but whether the injected history
provides motion-consistent evidence that can be stably used for action generation
(\autoref{fig:teaser}).

To understand this failure mode, we examine how 4D evidence is commonly
constructed. A typical discrete 4D injection pattern samples a finite set of
historical frames, lifts each selected frame into geometry-aware tokens through
RGB-D back-projection or learned 2D-derived geometry predictors~\cite{4DVLA,SwiftVLA},
and sparsifies the resulting history through memory or keyframe sampling
strategies~\cite{MemoryVLA,4DVLA}. Such a representation is 4D in form, but not
necessarily motion-consistent in time. Independent frame-wise lifting can introduce
\textbf{geometrical inconsistency}, where the same physical point is mapped to
drifting 3D locations across frames. Sparse or non-uniform history can further
cause \textbf{temporal fragmentation}, making the motion evidence irregular and
hard to exploit. These effects can propagate into action generation as jittery
corrections, unstable optimization, or detour-like executions.

We therefore argue that effective VLA memory should represent the motion that
connects observations, rather than merely storing more observations. To this end,
we introduce \textbf{MotionVLA}, a motion-history interface that converts a short
past-only observation window into compact, time-continuous trajectory-field
tokens. In our implementation, this interface is instantiated with trajectory-field
representations~\cite{4DGS, TraceAnything}, which provide continuous motion cues
over the recent history window. Instead of concatenating raw history or
independently lifted 4D features into the policy prefix, MotionVLA stores these
tokens as a queryable motion history. Current visual tokens then retrieve
task-relevant motion evidence from this history, and the retrieved information is
recoupled into the VLA stream under trajectory-grounded supervision. This design
keeps the role of motion history explicit: it should expose recent physical
progress to the policy, rather than simply enlarge the context with additional
frames or geometric tokens.

Experiments across RoboTwin2.0~\cite{RoboTwin2} and LIBERO~\cite{LIBERO}, together
with preliminary real-robot rollouts, show that MotionVLA improves long-horizon
manipulation while producing smoother and more direct executions. Ablations further
indicate that naive history aggregation or naive 4D token injection is not
sufficient; the gains come from aligning, querying, and recoupling temporally
consistent motion evidence so that it becomes usable for control.

Our main contributions are:
\begin{itemize}
    \item We propose \textbf{MotionVLA}, a compact time-continuous 
    motion-history interface that represents recent observations as 
    trajectory-field tokens rather than as independently lifted historical frames.

    \item We introduce a queryable motion history with a Decouple-then-Recouple 
    design: current visual tokens retrieve task-relevant motion evidence from the 
    past-only motion history before fusing it back into the VLA stream under 
    trajectory-grounded supervision.

    \item Experiments across RoboTwin2.0 and LIBERO, together with preliminary 
    real-robot rollouts, show that motion-consistent history improves 
    long-horizon manipulation and produces smoother, more direct executions.
\end{itemize}

\section{Related Work}

\paragraph{History and Memory in VLAs.}
Temporal modeling is central to long-horizon decision making in sequential domains such as autonomous driving~\cite{BEVFormer,UniAD,DiffusionDrive}.
In robot control, generalist VLAs transfer VLM priors to action generation and are often pretrained on heterogeneous robot-data mixtures~\cite{OXE,RDT1B,OpenVLA,pi0,Octo,CogACT,SmolVLA,SpatialVLA,CoTVLA,TraceVLA,pi05,OpenVLAOFT}.
Most policies nevertheless predict actions primarily from the current observation, which can fail under state ambiguity~\cite{MemoryVLA}.
Multi-frame conditioning~\cite{RT1} helps but increases context length; recent methods compress history through rendered trajectories~\cite{TraceVLA}, memory retrieval~\cite{MemoryVLA}, or compact temporal chunks~\cite{CronusVLA}.
Our work follows this memory-aware direction but uses a continuous trajectory-field interface rather than raw frame stacking or sparse keyframe selection.
Beyond temporal conditioning, complementary efforts improve controllability and representations, e.g., scaling action tokenizers~\cite{VQVLA} or injecting richer world knowledge into VLA training~\cite{DreamVLA}.
These directions improve the policy output space or semantic priors, while MotionVLA focuses on the consistency of historical evidence before it is exposed to the policy.

\paragraph{Injecting Geometry into VLAs.}
Geometry-aware policies incorporate metric cues such as depth maps, point clouds, or 3D embeddings from pretrained predictors~\cite{Dynamic2DGaussian,VGGT,DA3}.
DP3~\cite{DP3} shows the value of simple point-cloud inputs for visuomotor learning.
DepthVLA~\cite{DepthVLA} incorporates depth-aware spatial reasoning, while GeoVLA~\cite{GeoVLA} transforms depth into point clouds and fuses geometric embeddings with VLM features for action generation.
PointVLA~\cite{PointVLA} similarly explores point-cloud conditioning for VLA policies.
Spatially grounded VLMs~\cite{SpatialVLM,SpatialRGPT,VLM3R} also show that explicit geometry can improve language-conditioned perception.
These methods mainly address per-frame or static geometry; when such cues are applied independently across time, the temporal association between physical points can remain underspecified.
MotionVLA instead targets temporally consistent motion evidence.

\paragraph{4D Spatiotemporal Injection.}
Spatiotemporal VLMs and 4D representations have been explored for dynamic understanding and generation~\cite{VLM3R,VLM4D,Uni4DLLM,LangSplat4D}, and robot policies increasingly use space--time cues beyond static geometry.
Temporal modeling is also central in autonomous driving, where spatiotemporal BEV/3D stacks are widely used~\cite{DiffusionDrive,BEVFormer,StreamPETR,UniAD}.
ARM4R~\cite{ARM4R} lifts 2D tracking into 3D trajectories for robot pretraining; 4D-VLA~\cite{4DVLA} encodes sequential RGB-D observations with memory-bank sampling; SwiftVLA~\cite{SwiftVLA} derives auxiliary 4D features from streaming 2D images and a temporal cache.
These approaches provide 4D evidence as additional context, while our work is complementary: we focus on how the consistency of injected 4D evidence affects whether it can be reliably used for control.
This distinction matters for manipulation because small frame-to-frame geometric drift can manifest as inefficient corrections or end-effector detours even when a policy eventually completes the task.
This motivates using time-continuous trajectory fields~\cite{TrajectoryField, TraceAnything} as the motion-history interface.

\begin{figure}[t]
  \centering
  \includegraphics[max width=1\linewidth,trim=0.5cm 0cm 0.5cm 0cm,clip]{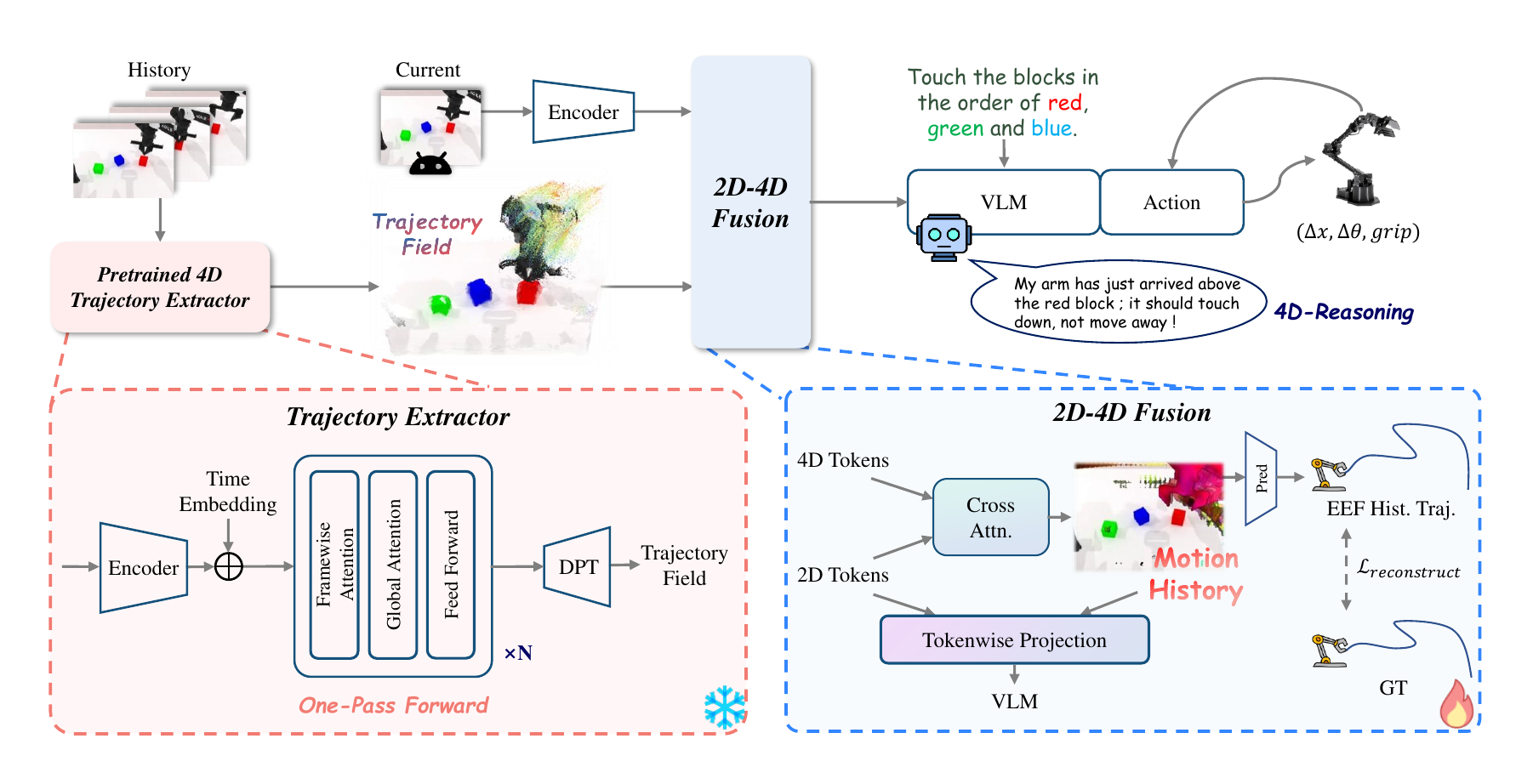}
  \caption{Overview of \textbf{MotionVLA}. MotionVLA builds a past-only motion history from trajectory-field tokens. Current visual features query this memory via cross-attention to retrieve task-relevant motion tokens, which are then recoupled into the VLA stream. An auxiliary trajectory reconstruction head (top right) grounds the retrieved tokens in control-relevant dynamics.
}
  \label{fig:overview}
\end{figure}

\section{Method}
\label{sec:method}

\subsection{Preliminaries}

\paragraph{VLA policy (\texorpdfstring{$\pi_0$}{pi0}).}
We study language-conditioned robot control at discrete timesteps $t$ \cite{pi0}.
The agent receives multi-view RGB observations $\mathbf{o}_t=\{I_t^{(k)}\}_{k=1}^{N_{\mathrm{cam}}}$ and proprioception $\mathbf{s}_t\in\mathbb{R}^{d_s}$, together with an instruction $x$.
A VLA policy outputs an $H$-step continuous action chunk $\mathbf{a}_t \in \mathbb{R}^{H\times d_a}$.
We write the tokenized perception-language-state prefix as
\begin{equation}
\mathbf{H}_t
=
\mathrm{LLM}\!\Big(
\big[
\mathrm{Enc}_{\mathrm{img}}(\mathbf{o}_t);\;
\mathrm{Enc}_{\mathrm{txt}}(x);\;
\mathrm{Tok}(\mathbf{s}_t)
\big]
\Big),
\end{equation}
where $[\cdot;\cdot]$ denotes token concatenation.
To model continuous actions, $\pi_0$ uses a flow-based action expert parameterized by a conditional velocity field $\mathbf{v}_{\theta}(\mathbf{a}^{\tau},\tau;\mathbf{H}_t)$, with $\tau\in[0,1]$, which is integrated from noise to generate $\mathbf{a}_t$ at inference.

\paragraph{Trajectory-field 4D representation.}
To summarize short-horizon dynamics, we use a frozen pretrained trajectory extractor (TraceAnything\cite{TraceAnything}) on a short head-camera sequence.
Given $T$ head-camera frames $\mathbf{I}^{\mathrm{head}}_{1:T}=\{I^{\mathrm{head}}_1,\ldots,I^{\mathrm{head}}_T\}$, the extractor represents each pixel by a continuous-time 3D trajectory function over normalized time,
$\mathcal{T}:(i,u,v)\mapsto\mathbf{x}_{i,u,v}(\cdot)\in C([0,1],\mathbb{R}^3)$,
where $i$ indexes the input frame and $(u,v)$ indexes pixel coordinates.
Besides dense trajectory-field outputs, the extractor also provides internal spatiotemporal tokens; we denote the final-layer decoder tokens by
$\mathbf{Z}=\mathrm{TrajEnc}_{\mathrm{tok}}(\mathbf{I}^{\mathrm{head}}_{1:T})$,
which serve as compact 4D features for downstream conditioning.

\subsection{Building a Past-only Motion History}
\label{sec:4d_tokens}

\paragraph{Past-only motion history.}
At timestep $t$, MotionVLA converts past head-camera observations into a queryable motion history $\mathbf{M}_{\mathrm{kv}}$.
The construction explicitly excludes the current frame $I_t^{\mathrm{head}}$ and uses the selected history window
$\mathcal{W}_t=
\left[
\tilde{I}^{\mathrm{head}}_{t-1-(T_{\mathrm{hist}}-1)\Delta},
\ldots,
\tilde{I}^{\mathrm{head}}_{t-1-\Delta},
\tilde{I}^{\mathrm{head}}_{t-1}
\right]$, where $\Delta$ is the temporal stride and $\tilde{I}^{\mathrm{head}}_{\tau}$ repeats the earliest available frame whenever $\tau$ precedes the episode start.
The window therefore spans roughly $T_{\mathrm{hist}}\Delta$ raw frames while remaining strictly past-only.
During training, $\mathcal{W}_t$ is obtained by indexing the same episode; during inference, the same definition is implemented online with a per-episode FIFO buffer, so train and test use an identical history interface.

\begin{wrapfigure}[14]{r}{0.55\linewidth}
\begin{minipage}{\linewidth}
\captionsetup{
    type=algorithm,
    font=small,
    labelfont=bf,
    textfont=bf,
    labelsep=space,
    justification=raggedright,
    singlelinecheck=false,
    skip=1pt
}
\hrule height 0.8pt
\vspace{0.5mm}
\captionof{algorithm}{Building the queryable motion history $\mathbf{M}_{\mathrm{kv}}$}
\label{alg:motion_history}
\vspace{0.5mm}
\hrule height 0.4pt
\vspace{0.5mm}
\footnotesize
\begin{algorithmic}[1]
\Require step $t$, history length $T_{\mathrm{hist}}$, stride $\Delta$
\Require episode frames or FIFO buffer $\mathcal{B}$
\If{training}
    \State $\mathcal{B} \gets [I^{\mathrm{head}}_1,\ldots,I^{\mathrm{head}}_{t-1}]$
\Else
    \State Append $I^{\mathrm{head}}_{t-1}$ to $\mathcal{B}$ and evict the oldest frame
\EndIf
\State Left-pad $\mathcal{B}$ with its earliest frame until
\Statex \hspace{\algorithmicindent} $|\mathcal{B}| \ge 1+(T_{\mathrm{hist}}-1)\Delta$
\State $L \gets |\mathcal{B}|$
\State $\mathcal{W}_t \gets \mathrm{StrideSample}(\mathcal{B},T_{\mathrm{hist}},\Delta)$
\State $\mathcal{W}_t \gets \mathrm{Norm}(\mathrm{Resize}(\mathcal{W}_t))$
\State $\mathbf{Z}_t \gets \mathrm{TrajEnc}_{\mathrm{tok}}(\mathcal{W}_t)$
\State $\mathbf{M}_{\mathrm{kv}}\gets\mathrm{PosEnc}(W_z\mathbf{Z}_t)\in\mathbb{R}^{B\times S_{\mathrm{4D}}\times E}$
\State \textbf{return} $\mathbf{M}_{\mathrm{kv}}$
\end{algorithmic}
\vspace{0.5mm}
\hrule height 0.8pt
\end{minipage}
\vspace{4mm}
\end{wrapfigure}

\paragraph{Compact token interface.}
The output of Algorithm~\ref{alg:motion_history} is the queryable motion history $\mathbf{M}_{\mathrm{kv}}$.
We do not feed dense trajectory fields to the VLA directly; instead, we keep only the final decoder tokens and project them to the VLA hidden size through $W_z$.
This token-level interface preserves cross-frame motion evidence while avoiding the token explosion and local tracking-noise sensitivity of dense trajectory-field outputs.
Crucially, $\mathbf{M}_{\mathrm{kv}}$ is used only as a key--value memory for later retrieval, so the current observation still determines which parts of the past are read, keeping motion history as a compact control-relevant memory.

\subsection{Decouple-then-Recouple 4D Fusion}
\begingroup
\setlength{\abovedisplayskip}{3pt}
\setlength{\belowdisplayskip}{3pt}
\setlength{\abovedisplayshortskip}{2pt}
\setlength{\belowdisplayshortskip}{2pt}

Our design follows a simple chain:
\textbf{(i) extract} 4D evidence with a trajectory extractor,
\textbf{(ii) decouple} task-relevant motion via query-conditioned retrieval,
\textbf{(iii) supervise} the retrieved tokens to preserve temporal grounding, and
\textbf{(iv) recouple} motion with current perception for action generation.

\paragraph{Decouple: query-conditioned retrieval.}
Let $\mathbf{V}^{\mathrm{head}}\in\mathbb{R}^{B\times S_{\mathrm{img}}\times E}$ denote the current head-camera visual tokens produced by the VLA visual encoder (\autoref{fig:overview}).
We use these current-frame tokens as queries to retrieve motion-conditioned evidence from the motion history:
\begin{equation}
\mathbf{M}
=
\mathrm{MHA}\!\left(
\mathbf{V}^{\mathrm{head}},
\mathrm{LN}(\mathbf{M}_{\mathrm{kv}}),
\mathrm{LN}(\mathbf{M}_{\mathrm{kv}})
\right),
\label{eq:motiontokens}
\end{equation}
where $\mathbf{M}\in\mathbb{R}^{B\times S_{\mathrm{img}}\times E}$.
Thus, $\mathbf{M}$ has the same token length as the current head-view queries, but each token is conditioned on the longer past-only motion history.
This query-conditioned retrieval aligns historical motion evidence to the current observation, rather than naively concatenating all historical features into the policy prefix.

\paragraph{Trajectory-grounded motion tokens.}
Query-conditioned retrieval alone may still allow motion-conditioned tokens $\mathbf{M}$ to collapse into appearance summaries; during training we therefore optimize the joint objective
$\mathcal{L}=\mathcal{L}_{\mathrm{action}}+\alpha\,\mathcal{L}_{\mathrm{traj}}$.
For the action term, we follow $\pi_0$'s flow-matching behavior cloning:
given an expert action chunk $\mathbf{a}\in\mathbb{R}^{H\times d_a}$, we sample $\tau\sim\mathcal{U}(0,1)$ and $\boldsymbol{\epsilon}\sim\mathcal{N}(\mathbf{0},\mathbf{I})$, construct $\mathbf{a}^{\tau}=(1-\tau)\boldsymbol{\epsilon}+\tau\mathbf{a}$, and regress the policy's predicted velocity field $\mathbf{v}_{\theta}(\mathbf{a}^{\tau},\tau;\text{prefix})$ to the target field $\mathbf{u}(\mathbf{a}^{\tau}\mid\mathbf{a})=\mathbf{a}-\boldsymbol{\epsilon}$:
\begin{equation}
\mathcal{L}_{\mathrm{action}}
=
\mathbb{E}\!\left[
\left\|
\mathbf{v}_{\theta}(\mathbf{a}^{\tau},\tau;\text{prefix})
-
(\mathbf{a}-\boldsymbol{\epsilon})
\right\|_2^2
\right].
\label{eq:laction}
\end{equation}

To ground $\mathbf{M}$ in temporally meaningful progress (\autoref{fig:overview}), we add an auxiliary history-action reconstruction objective.
Let $\mathbf{a}_{\mathrm{hist}}\in\mathbb{R}^{T_{\mathrm{hist}}\times d_a}$ denote the action sequence aligned with the same history window used to build the motion history.
We attach a lightweight prediction head $g_{\phi}$, predict $\hat{\mathbf{a}}_{\mathrm{hist}}=g_{\phi}(\mathbf{M})$, and minimize
\begin{equation}
\mathcal{L}_{\mathrm{traj}}
=
\frac{1}{T_{\mathrm{hist}}d_a}
\left\|
\hat{\mathbf{a}}_{\mathrm{hist}}-\mathbf{a}_{\mathrm{hist}}
\right\|_2^2.
\label{eq:ltraj}
\end{equation}
This auxiliary supervision keeps the main action objective unchanged while encouraging the retrieved motion-conditioned tokens to encode control-relevant dynamics rather than static appearance differences.
We use action sequences, instead of reconstructing pixels or dense trajectories directly, because they more directly reflect the robot's dynamical progress and state evolution.

\paragraph{Recouple: lightweight fusion into the VLA stream.}
We then recouple the motion-conditioned current tokens with the VLA's current multi-view visual tokens using a lightweight fusion module (\autoref{fig:overview}).
Let $\mathbf{V}^{\mathrm{mv}}\in\mathbb{R}^{B\times S_{\mathrm{mv}}\times E}$ denote the multi-view visual tokens at time $t$.
We concatenate motion-conditioned and multi-view visual tokens and apply a token-mixing MLP to obtain fused tokens
\[
\mathbf{F}
=
\mathrm{MLP}\!\left([\mathbf{M};\mathbf{V}^{\mathrm{mv}}]\right)
\in
\mathbb{R}^{B\times (S_{\mathrm{img}}+S_{\mathrm{mv}})\times E}.
\]
Finally, we feed $\mathbf{F}$ to the VLA backbone together with text tokens and robot-state tokens, enabling the flow-based action expert to condition on both current perception and retrieved motion evidence through shared self-attention.

\endgroup
\FloatBarrier
\providecommand{\tabyes}{\ensuremath{\checkmark}}
\providecommand{\tabno}{\ensuremath{\times}}

\section{Experiments}
\label{sec:experiments}

We organize the evaluation around three questions aligned with our main claims:
(i) does motion history improve long-horizon manipulation success,
(ii) does it produce more motion-consistent executions rather than merely higher terminal success,
and (iii) are the gains due to the proposed alignment, retrieval, and recoupling design rather than simply adding more history or 4D tokens?

\subsection{Experimental Protocol}
\label{sec:exp_protocol}

\paragraph{Benchmarks.}
We evaluate MotionVLA on RoboTwin2.0~\cite{RoboTwin2} and LIBERO~\cite{LIBERO}.
For RoboTwin2.0, we use six manipulation tasks grouped by episode length:
two long-horizon tasks ($>$400 steps), two mid-horizon tasks (300--400 steps), and two short-horizon tasks ($<$300 steps).
One long-horizon task, \emph{blocks\_touching\_rgb}, is a custom RoboTwin2.0 task designed to stress multi-stage temporal reasoning; the other five tasks are standard RoboTwin2.0 tasks.
Detailed task definitions are provided in Appendix~\ref{app:robotwin_tasks}.
For LIBERO, we evaluate on \textsc{LIBERO-Spatial}, \textsc{LIBERO-Object}, \textsc{LIBERO-Goal}, and \textsc{LIBERO-Long} following the standard protocol, and report suite-wise success rates and their average.

\paragraph{Metrics.}
We report success rate (SR) for all benchmarks.
To evaluate the motion-consistency claim beyond terminal success, we also report path efficiency,
defined as the ratio between the end-effector travel length executed by a policy and the corresponding expert demonstration length.
For an episode with end-effector positions $\{p_t\}_{t=1}^{T_{\mathrm{episode}}}$, we compute
\begin{equation}
L=\sum_{t=1}^{T_{\mathrm{episode}}-1}\|p_{t+1}-p_t\|_2 ,
\end{equation}
and define
\begin{equation}
\mathrm{PE}=\frac{L_{\mathrm{policy}}}{L_{\mathrm{expert}}}.
\end{equation}
We average PE over successful episodes and always report it together with SR, since PE alone can be biased when different methods succeed on different subsets of episodes.
Lower PE indicates fewer detours, with $\mathrm{PE}=1$ matching expert path length.

\paragraph{Implementation details.}
Unless otherwise specified, the head-camera stream is sampled at 10 FPS, and MotionVLA uses $T_{\mathrm{hist}}=5$ selected past frames with stride $\Delta=2$, covering 10 raw camera frames.
Training and inference use the same past-only history definition: the current frame is excluded, and if the available past context is too short to fill the selected strided history, the window is left-padded by repeating the earliest available frame.
The trajectory extractor is frozen and run once per timestep on the subsampled head-camera history.
Training follows a generic-to-specific recipe.
\emph{Stage I} trains on a heterogeneous manipulation mixture (100 hours) to align the motion-history interface with the VLA embedding space using the action loss and $\mathcal{L}_{\mathrm{traj}}$.
\emph{Stage II} adapts the policy to target tasks with the auxiliary trajectory head detached.
Unless otherwise specified, both stages use learning rate $1\times10^{-5}$ and batch size 32.
For preliminary real-world validation on Agilex Piper, we collect more than 100 demonstrations per task, train the model on 8 NVIDIA H20 GPUs, and evaluate it on a single NVIDIA A800 GPU.

\subsection{Main Results}
\label{sec:main_results}

\subsubsection{Simulation Benchmarks}
\label{sec:sim_results}

\begin{table*}[t]
\centering
\caption{Main results on RoboTwin2.0 (success rate \%). Tasks are grouped by average episode length: long-term ($>$400 steps), mid-term (300--400), and short-term ($<$300). ``Touch'' is \emph{blocks\_touching\_rgb}, ``Rank'' is \emph{blocks\_ranking\_rgb}, ``Stack2'' is \emph{stack\_blocks\_two}, ``Stack3'' is \emph{stack\_bowls\_three}, ``Place'' is \emph{place\_a2b\_right}, ``Hand'' is \emph{handover\_block}.}
\label{tab:robotwin_main}
\vspace{-2mm}
\setlength{\tabcolsep}{2.6pt}
\renewcommand{\arraystretch}{1.05}
\resizebox{\textwidth}{!}{%
\begin{tabular}{l|c||>{\centering\arraybackslash}m{1.2cm}
                    >{\centering\arraybackslash}m{1.2cm}|
                    >{\centering\arraybackslash}m{1.2cm}
                    >{\centering\arraybackslash}m{1.2cm}|
                    >{\centering\arraybackslash}m{1.2cm}
                    >{\centering\arraybackslash}m{1.2cm}
                    |c}
\toprule
\rowcolor[gray]{.9}
 &&\multicolumn{2}{c|}{Long-Term} &
\multicolumn{2}{c|}{Mid-Term} &
\multicolumn{2}{c|}{Short-Term} &
 \\
\rowcolor[gray]{.9}
\multirow{-2}{*}{Method}&\multirow{-2}{*}{Publication} &
Touch  & Rank  
& Stack2 & Stack3
& Place  & Hand  
& \multirow{-2}{*}{Avg.}\\
\midrule
DP \cite{DP}                    & RSS'23 & 0  & 0  & 7  & 63 & 13 & 10 & 16 \\
DP3 \cite{DP3}                  & arXiv'24   & 0  & 3  & 24 & 57 & \textbf{49} & \textbf{70} & 34 \\
RDT \cite{RDT1B}                & ICLR'25    & 0  & 1  & 25 & 48 & 0  & 42 & 19 \\
OpenVLA-OFT \cite{OpenVLAOFT}   & arXiv'25    & 3  & 45 & 53 & 62 & 23 & 11 & 33 \\
\midrule
$\pi_0$ (baseline) \cite{pi0}       & RSS'25 & 19 & 43 & 43 & 55 & 35 & 48 & 41 \\
Ours (w/o Recouple)            & ---    & 37 & \textbf{53} & 52 & 63 & 45 & 55 & 51 \\
\rowcolor{cyan!10}Ours (w/ Recouple)             & ---    & \textbf{41} & \textbf{53} & \textbf{57} & \textbf{67} & 37 & 63 & \textbf{53} \\
\bottomrule
\end{tabular}
}
\vspace{-2mm}
\end{table*}

\begin{table*}[t]
\vspace{1mm}
\centering
\caption{Results on LIBERO (success rate \%). \textbf{History} indicates whether the policy explicitly conditions on past observation frames at inference time (multi-view inputs at the same timestep are marked as \xmark). \textbf{3D} indicates whether the policy explicitly uses 3D cues at inference time (e.g., depth/point clouds/3D coordinate embeddings).}
\label{tab:libero_main}
\vspace{-2mm}
\setlength{\tabcolsep}{4.0pt}
\renewcommand{\arraystretch}{1.05}
\begin{tabular}{l |c || c c c c c c| c}
\toprule
\rowcolor[gray]{.9}
Method & Publication & History & 3D & Spatial & Object & Goal & Long & Avg. \\
\midrule
Diffusion Policy \cite{DP}      & RSS'23     & \xmark & \xmark & 78.3 & 92.5 & 68.3 & 50.5 & 72.4 \\
Octo \cite{Octo}               & RSS'24     & \xmark & \xmark & 78.9 & 85.7 & 84.6 & 51.1 & 75.1 \\
OpenVLA \cite{OpenVLA}         & CoRL'24    & \xmark & \xmark & 84.7 & 88.4 & 79.2 & 53.7 & 76.5 \\
CoT-VLA \cite{CoTVLA}          & CVPR'25    & \xmark & \xmark & 87.5 & 91.6 & 87.6 & 69.0 & 83.9 \\
$\pi_0$-FAST \cite{pi0FAST}            & arXiv'25     & \xmark & \xmark & 96.4 & 96.8 & 88.6 & 60.2 & 85.5 \\
PixelVLA \cite{PixelVLA}       & arXiv'25   & \xmark & \xmark & 88.5 & 90.0 & 85.8 & 82.6 & 86.7 \\
GR00T-N1 \cite{GR00T}          & arXiv'25   & \xmark & \xmark & 94.4 & 97.6 & 93.0 & \underline{90.6} & 93.9 \\
$\pi_0$ \cite{pi0}                 & RSS'25     & \xmark & \xmark & \underline{96.8} & \textbf{98.8} & 95.8 & 85.2 & 94.2 \\
\midrule
TraceVLA \cite{TraceVLA}       & ICLR'25    & \cmark & \xmark & 84.6 & 85.2 & 75.1 & 54.1 & 74.8 \\
SpatialVLA \cite{SpatialVLA}   & RSS'25     & \xmark & \cmark & 88.2 & 89.9 & 78.6 & 55.5 & 78.1 \\
DepthVLA \cite{DepthVLA}       & arXiv'25   & \xmark & \cmark & 96.4 & \underline{98.0} & 95.8 & 89.2 & 94.9 \\
\midrule
4D-VLA \cite{4DVLA}            & NeurIPS'25 & \cmark & \cmark & 88.9 & 95.2 & 90.9 & 79.1 & 88.6 \\
SwiftVLA \cite{SwiftVLA}       & arXiv'25   & \cmark & \cmark & \textbf{97.2} & 96.8 & \textbf{97.4} & 89.0 & \underline{95.1} \\
\cellcolor{cyan!10}Ours        & \cellcolor{cyan!10}--- &
\cellcolor{cyan!10}\cmark & \cellcolor{cyan!10}\cmark &
\cellcolor{cyan!10}96.2 & \cellcolor{cyan!10}\underline{98.0} & \cellcolor{cyan!10}\underline{96.2} &
\cellcolor{cyan!10}\textbf{91.2} & \cellcolor{cyan!10}\textbf{95.4} \\
\bottomrule
\end{tabular}
\vspace{-2mm}
\end{table*}

Tables~\ref{tab:robotwin_main} and~\ref{tab:libero_main} summarize the simulation results.
On RoboTwin2.0, MotionVLA reaches $53\%$ average success, improving over $\pi_0$ by 12 points; on the long-horizon Touch task, success increases from $19\%$ to $41\%$.
On LIBERO, MotionVLA obtains the best average score among the listed methods ($95.4\%$) and the best \textsc{LIBERO-Long} score ($91.2\%$), improving over $\pi_0$ by 6.0 points on the long-horizon suite.
Together with the weaker history-only and geometry-only baselines, these results suggest that the gain comes from aligned, queryable motion history rather than generic extra context.

\subsubsection{Real-world Validation}
\label{sec:real_robot_main}

\begin{table}[t]
\centering
\caption{\textbf{Real-world validation on Agilex Piper.} SR is success rate (\%); steps are computed by multiplying completion time by 10 and rounding to the nearest integer. For Avg. Steps, unavailable task-level step values are excluded from the average.}
\vspace{1mm}
\label{tab:real_robot}
\setlength{\tabcolsep}{5.0pt}
\renewcommand{\arraystretch}{1.2}
\scriptsize
\begin{tabular}{l|cc|cc|cc|cc}
\toprule
\rowcolor[gray]{.9} 
& \multicolumn{2}{c|}{Pick\&Place} 
& \multicolumn{2}{c|}{Ranking} 
& \multicolumn{2}{c|}{Touching} 
& \multicolumn{2}{c}{Avg.} \\
\rowcolor[gray]{.9} 
Method 
& SR$\uparrow$ & Steps$\downarrow$ 
& SR$\uparrow$ & Steps$\downarrow$ 
& SR$\uparrow$ & Steps$\downarrow$ 
& SR$\uparrow$ & Steps$\downarrow$ \\
\midrule
Demo 
& -- & 119 
& -- & 351 
& -- & 261 
& -- & 244 \\
$\pi_0$ 
& 80.0 & 118 
& 6.1 & 558 
& 0.0 & -- 
& 28.7 & 338 \\
\rowcolor{cyan!10} 
Ours 
& \textbf{92.5} & \textbf{119} 
& \textbf{18.2} & \textbf{443} 
& \textbf{15.0} & \textbf{285} 
& \textbf{41.9} & \textbf{282} \\
\bottomrule
\end{tabular}
\vspace{-4mm}
\end{table}

On an Agilex Piper setup with three temporally demanding tasks (Table~\ref{tab:real_robot}), MotionVLA improves average success from $28.7\%$ to $41.9\%$ and reduces average completion steps from 338 to 282.
The improvement is visible not only on Pick\&Place but also on the more temporally structured Ranking and Touching tasks.
These results provide preliminary real-world evidence for the motion-history interface, but are not intended as a comprehensive deployment study across embodiments or environments.
Qualitative rollouts and trajectory-field visualizations are provided in Appendix~\ref{app:qualitative}.

\subsection{Motion Consistency and Component Evidence}
\label{sec:motion_and_ablation}

\begin{wraptable}[18]{r}{0.4\linewidth}
\vspace{-4mm}
\centering
\captionsetup{type=table}

\begin{minipage}{\linewidth}
\centering
\caption{\textbf{Path efficiency on LIBERO.}}
\label{tab:path_efficiency}
\vspace{-1mm}
\setlength{\tabcolsep}{2.0pt}
\renewcommand{\arraystretch}{0.96}
\footnotesize
\begin{tabular}{l|cc|cc}
\toprule
\rowcolor[gray]{.9}
& \multicolumn{2}{c|}{\textsc{Goal}}
& \multicolumn{2}{c}{\textsc{Long}} \\
\rowcolor[gray]{.9}
Method & SR$\uparrow$ & PE$\downarrow$ & SR$\uparrow$ & PE$\downarrow$ \\
\midrule
$\pi_0$~\cite{pi0}       & 95.8 & 1.23 & 85.2 & 1.19 \\
4D-VLA*~\cite{4DVLA}     & 88.8 & 1.88 & 72.6 & 2.06 \\
4D-VLA~\cite{4DVLA}      & 89.8 & 1.36 & 78.6 & 1.32 \\
\rowcolor{cyan!10}
Ours                     & \textbf{96.2} & \textbf{1.05} & \textbf{91.2} & \textbf{1.10} \\
\bottomrule
\end{tabular}
\end{minipage}

\vspace{5mm}

\begin{minipage}{\linewidth}
\centering
\caption{\textbf{Component ablation.} T1/T2 are Touch/Stack3; Stg1/Aux/Proj denote Stage-I, auxiliary loss, and projector.}
\label{tab:component_ablation}
\vspace{-1mm}
\setlength{\tabcolsep}{1.5pt}
\renewcommand{\arraystretch}{0.96}
\footnotesize
\begin{tabular}{ccccc||cc|c}
\toprule
\rowcolor{gray!10}
Hist. & 4D & Stg1 & Aux & Proj & T1 & T2 & Avg. \\
\midrule
$\times$ & $\times$ & $\times$ & $\times$ & $\times$ & 19 & 55 & 37 \\
\midrule
\checkmark & $\times$ & $\times$ & $\times$ & $\times$ & 3  & 23 & 13 \\
\checkmark & \checkmark & $\times$ & $\times$ & $\times$ & 12 & 19 & 16 \\
\checkmark & \checkmark & \checkmark & $\times$ & $\times$ & 34 & 45 & 40 \\
\checkmark & \checkmark & \checkmark & \checkmark & $\times$ & 37 & 63 & 50 \\
\rowcolor{cyan!10}
\checkmark & \checkmark & \checkmark & \checkmark & \checkmark & \textbf{41} & \textbf{67} & \textbf{54} \\
\bottomrule
\end{tabular}
\end{minipage}

\vspace{-3mm}
\end{wraptable}

\paragraph{Motion consistency.}
Table~\ref{tab:path_efficiency} tests whether higher success also yields more direct executions.
On \textsc{LIBERO-Goal}/\textsc{LIBERO-Long}, MotionVLA obtains PE $=1.05/1.10$, closer to expert path length than both $\pi_0$ ($1.23/1.19$) and 4D-VLA ($1.36/1.32$).
The pseudo-RGBD 4D-VLA* variant is less efficient, indicating that noisy per-frame geometry can introduce detours rather than improve motion consistency.

\paragraph{Component evidence.}
Table~\ref{tab:component_ablation} rules out a simple ``more history or more tokens'' explanation.
Naive visual history reduces the two-task average from $37\%$ to $13\%$, and unaligned 4D tokens reach only $16\%$.
Stage-I alignment raises the average to $40\%$, while auxiliary trajectory grounding and recoupling further improve it to $50\%$ and $54\%$.
This pattern supports the decouple-then-recouple interface: motion history must be aligned, queried by the current observation, and reinjected into the VLA token space.
\section{Conclusion}
\label{sec:conclusion}
We introduced \textbf{MotionVLA}, a time-continuous motion-history interface for $\pi_0$-style VLA policies that recouples compact trajectory-field tokens with current perception through trajectory-grounded supervision.
Across evaluated simulation settings, MotionVLA improves long-horizon consistency and path efficiency over the baseline; preliminary Agilex Piper results show the same trend in success and speed.
These results support the central premise that resolving spatiotemporal inconsistency in the injected evidence is as important as adding more historical observations.
Thus, MotionVLA changes history from fragmented frame evidence into a queryable motion representation that better matches manipulation geometry.
Future work includes robust trajectory extraction, broader history schedules, and larger-scale real-world validation.

\section{Limitations}
\label{sec:Limitations}
Motion history is most useful under temporal ambiguity or long-horizon state confusion, but can be neutral or slightly harmful when the current observation is sufficient.
The frozen trajectory extractor can fail under occlusion, fast motion, or textureless scenes; broader history-size search and stronger-backbone integration remain future work.
Our real-world results are preliminary and cover few Agilex Piper tasks, so they should be read as evidence that the interface transfers beyond simulation rather than as a comprehensive deployment study.


\bibliography{Paper-4D-Pi}

\clearpage
\clearpage
\appendix
\onecolumn
\raggedbottom
\setcounter{page}{1}
\renewcommand{\thepage}{A\arabic{page}}
\providecommand{\theHpage}{\thepage}
\renewcommand{\theHpage}{appendix.\arabic{page}}
\section*{Appendix}

\section{Additional Experimental Results and Ablations}
\label{app:additional_exp}

\subsection{Training Efficiency and Stage-I Sensitivity}
\label{app:efficiency_stage1}

\begin{figure}[H]
  \centering
  \begin{minipage}[t]{0.43\linewidth}
    \centering
    \includegraphics[width=\linewidth]{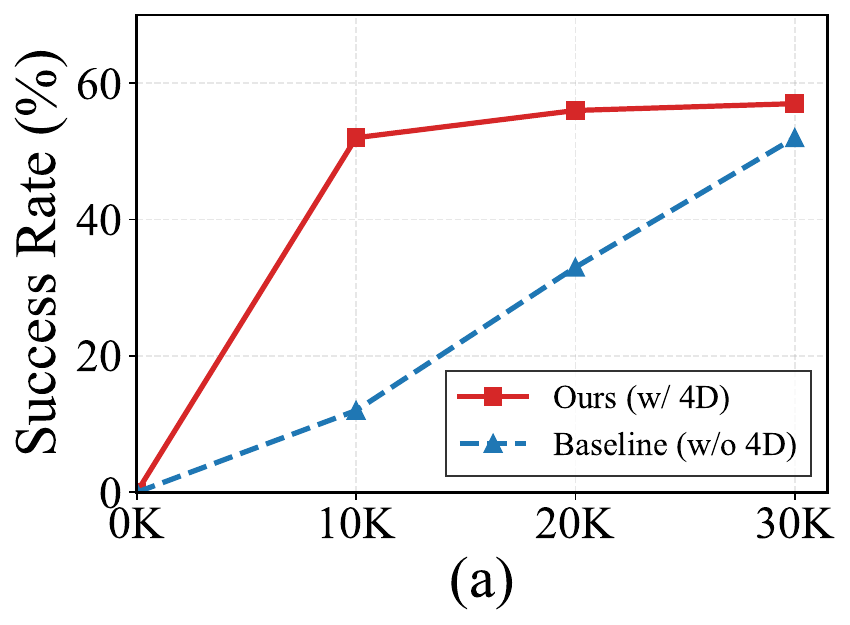}\\[-1mm]
    {\footnotesize (a) Training efficiency.}
  \end{minipage}
  \hspace{0.06\linewidth}
  \begin{minipage}[t]{0.43\linewidth}
    \centering
    \includegraphics[width=\linewidth]{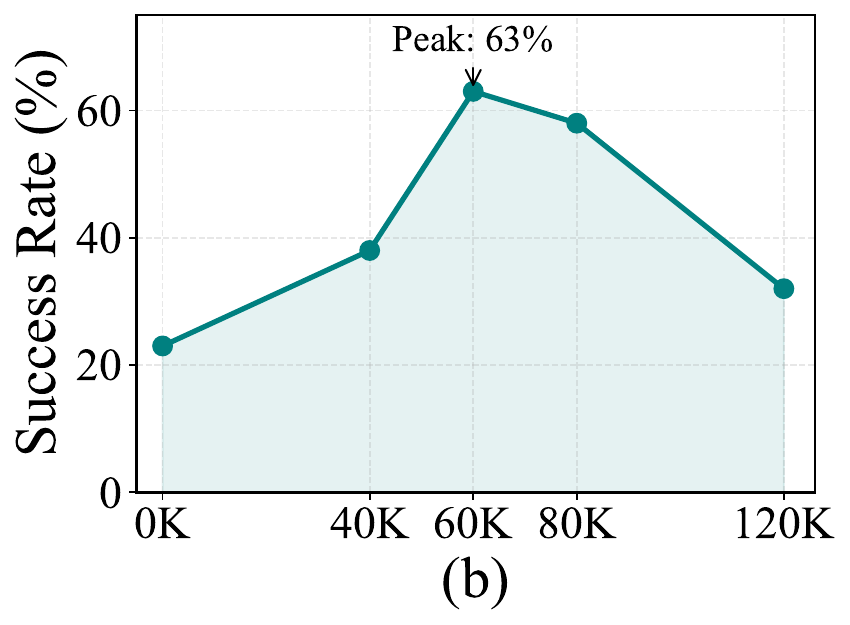}\\[-1mm]
    {\footnotesize (b) Stage-I training steps.}
  \end{minipage}
  \vspace{-1mm}
  \caption{\textbf{Training efficiency and Stage-I sensitivity.}
  On \emph{stack\_blocks\_two}, MotionVLA reaches higher early success than the baseline, suggesting that motion-history tokens provide a useful optimization prior.
  Stage-I alignment peaks around 60k steps in this setting; insufficient alignment underuses trajectory-field features, while overly long Stage-I training can slightly hurt downstream adaptation.}
  \label{fig:app_efficiency_and_stage1}
  \vspace{-2mm}
\end{figure}

MotionVLA converges faster during downstream adaptation because the motion-history interface reduces ambiguity in mapping visually similar intermediate states to actions.
The Stage-I sweep further suggests that the trajectory-field interface should be aligned to the VLA embedding space, but not over-specialized to the heterogeneous pretraining mixture before target-task adaptation.

\subsection{History Window, Stride, and Inference Speed}
\label{app:history_window}

\begin{table}[H]
\centering
\caption{\textbf{History-window ablation} on \emph{blocks\_touching\_rgb}. Eff. denotes the effective raw-frame temporal coverage \(T_{\mathrm{hist}}\Delta\).}
\label{tab:app_history_ablation}
\vspace{-1mm}
\setlength{\tabcolsep}{7.0pt}
\renewcommand{\arraystretch}{1.08}
\small
\begin{tabular}{cccc|c}
\toprule
\rowcolor[gray]{.92}
\(T_{\mathrm{hist}}\) & \(\Delta\) & Eff. & FPS & SR \\
\midrule
0  & -- & -- & 12.3 & 19 \\
3  & 2  & 6  & 8.9  & 33 \\
5  & 1  & 5  & 7.3  & 35 \\
\rowcolor{cyan!10}
5  & 2  & 10 & 7.1  & \textbf{41} \\
10 & 1  & 10 & 3.1  & 40 \\
10 & 2  & 20 & 3.1  & 23 \\
20 & 2  & 40 & 0.9  & 26 \\
\bottomrule
\end{tabular}
\vspace{-2mm}
\end{table}

The default setting \(T_{\mathrm{hist}}=5,\Delta=2\) gives the best observed success--speed trade-off in this ablation.
Compared with \(T_{\mathrm{hist}}=5,\Delta=1\), increasing the stride improves success from \(35\%\) to \(41\%\) while preserving similar inference speed.
However, longer effective histories do not consistently help: they increase computation and may introduce stale or task-irrelevant early-stage motion, making retrieval less focused for the current control state.

\subsection{Motion-history Alignment Diagnostics}
\label{app:alignment_diagnostics}

\begin{figure}[H]
  \centering
  \includegraphics[width=0.99\linewidth]{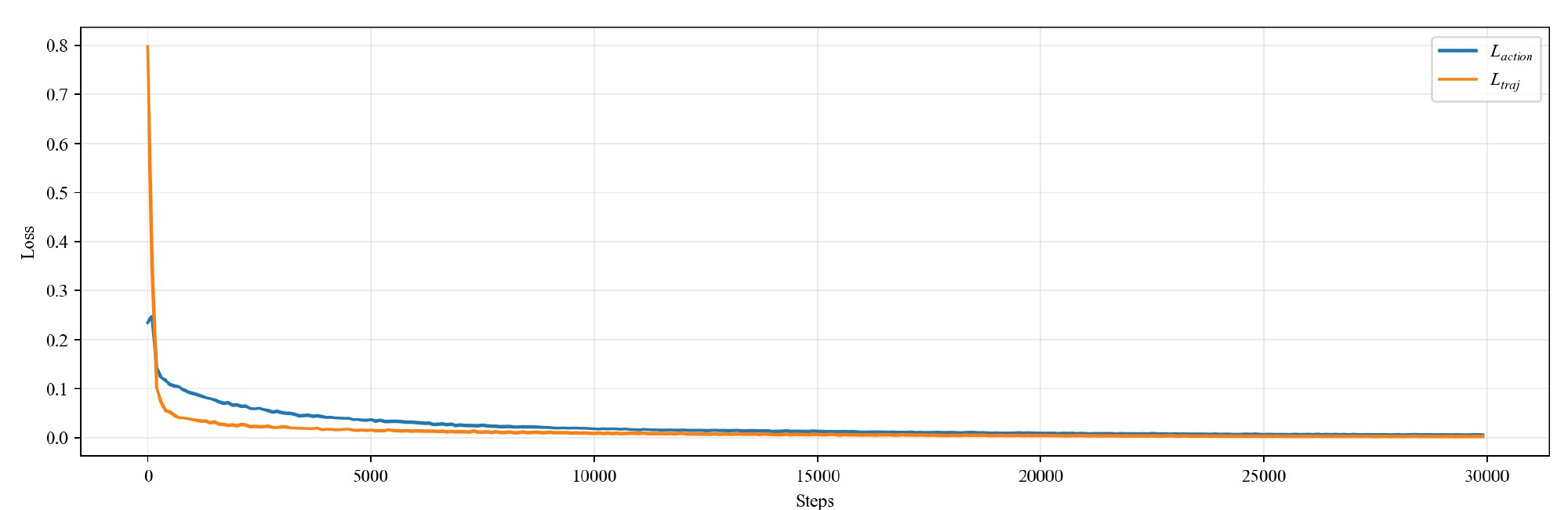}
  \vspace{-1mm}
  \caption{\textbf{Training curves during motion-history interface alignment.}
  We plot the action loss and the auxiliary trajectory-grounding loss used in Stage-I alignment.}
  \label{fig:app_loss_curves}
  \vspace{-2mm}
\end{figure}

The auxiliary trajectory-grounding loss is used only to shape the retrieved motion-conditioned tokens during alignment.
In Stage II, the auxiliary trajectory head is detached, and the downstream policy is optimized for action generation.
This separation keeps the motion-history tokens grounded in recent robot dynamics while avoiding an additional inference-time prediction requirement.

\subsection{Controlled Robustness to Motion-History Corruption}
\label{app:motion_history_corruption}

The frozen trajectory extractor is a potential source of error under
occlusion, fast motion, appearance perturbation, or textureless regions.
To quantify how such errors affect MotionVLA, we conduct a controlled
robustness evaluation on RoboTwin2.0. For each task, we load the
MotionVLA checkpoint trained with clean inputs. At test time, we corrupt
only the past head-camera frames \(W_t\) that are fed to the frozen
trajectory extractor, while keeping the current observation, proprioception,
and language instruction unchanged. This intervention isolates the
sensitivity of the motion-history branch from the main current-observation
pathway.

We apply three increasing corruption levels: mild, medium, and severe.
Each level combines random spatial occlusion, frame-wise appearance
perturbation, and random dropping of history frames. We use this setting as
a controlled proxy for trajectory-extractor failure rather than as a complete
robustness benchmark for all possible visual corruptions. Since the
\(\pi_0\) baseline does not use the motion-history branch, we report its
clean performance as a reference.

\begin{table}[H]
\centering
\small
\caption{
Controlled robustness analysis on RoboTwin2.0.
We corrupt only the history frames fed to the frozen trajectory extractor at
test time, while keeping the current observation unchanged. SR is reported
in percentage. ``Rank'' denotes \texttt{blocks\_ranking\_rgb}, and ``Hand''
denotes \texttt{handover\_block}. Avg. is the mean over the two tasks.
}
\label{tab:motion_history_corruption}
\begin{tabular}{llccc}
\toprule
Model & Corruption & Rank & Hand & Avg. \\
\midrule
\(\pi_0\) & Clean  & 43 & 48 & 45.5 \\
\midrule
MotionVLA & Clean  & 53 & 63 & 58.0 \\
MotionVLA & Mild   & 53 & 62 & 57.5 \\
MotionVLA & Medium & 50 & 59 & 54.5 \\
MotionVLA & Severe & 45 & 53 & 49.0 \\
\bottomrule
\end{tabular}
\end{table}

As shown in Table~\ref{tab:motion_history_corruption}, MotionVLA degrades
progressively rather than collapsing abruptly when the history branch is
corrupted. Compared with clean MotionVLA, the average success rate drops by
0.5, 3.5, and 9.0 points under mild, medium, and severe corruption,
respectively. Even under severe corruption, MotionVLA remains above the
clean \(\pi_0\) baseline on both tasks. These results support a bounded
robustness claim under this controlled history-branch corruption setting,
while also confirming that severe trajectory-extractor failures can still
degrade action quality.

\clearpage
\section{Task Definitions in RoboTwin2.0}
\label{app:robotwin_tasks}

\begingroup\small
We evaluate MotionVLA on six RoboTwin2.0 manipulation tasks (Table~\ref{tab:robotwin_main}). 
Among them, \emph{blocks\_touching\_rgb} is a custom task constructed in the RoboTwin2.0 simulation environment, while the other five tasks are standard RoboTwin2.0 tasks. 
Below we describe the goal and the success criteria for each task.

\paragraph{blocks\_touching\_rgb (custom).}
\textbf{Goal.} The robot uses the gripper tip to touch three colored blocks in the order \textbf{red} $\rightarrow$ \textbf{green} $\rightarrow$ \textbf{blue}. 
\textbf{Success.} The episode is successful if the gripper tip makes physical contact with the red, green, and blue blocks exactly once each in the correct order, without repeated touches or missed touches. 

\paragraph{blocks\_ranking\_rgb.}
\textbf{Goal.} The robot picks up three blocks and arranges them into a left-to-right line with the color order \textbf{red} $\rightarrow$ \textbf{green} $\rightarrow$ \textbf{blue}. 
\textbf{Success.} The episode is successful if the three blocks are placed approximately on a straight line and the left-to-right color ordering matches the target ranking. 

\paragraph{stack\_blocks\_two.}
\textbf{Goal.} The robot grasps the green block and stacks it on top of the red block. 
\textbf{Success.} The episode is successful if the green block stably remains above the red block (i.e., a valid two-block stack is formed and maintained). 

\paragraph{stack\_bowls\_three.}
\textbf{Goal.} The robot stacks three bowls into a single three-bowl stack. 
\textbf{Success.} The episode is successful if all three bowls are stacked together. 

\paragraph{place\_a2b\_right.}
\textbf{Goal.} The robot places object A (randomly sampled) to the right of object B (randomly sampled). 
\textbf{Success.} The episode is successful if object A is positioned to the right of object B. 

\paragraph{handover\_block.}
\textbf{Goal.} A bimanual task: the left arm grasps a red cube, hands it over to the right arm, and the right arm places the cube onto a blue pad. 
\textbf{Success.} The episode is successful if the red cube stably remains on the blue pad at the end of the episode. 
\endgroup

\begin{figure}[H]
    \centering
    \includegraphics[width=0.9\linewidth]{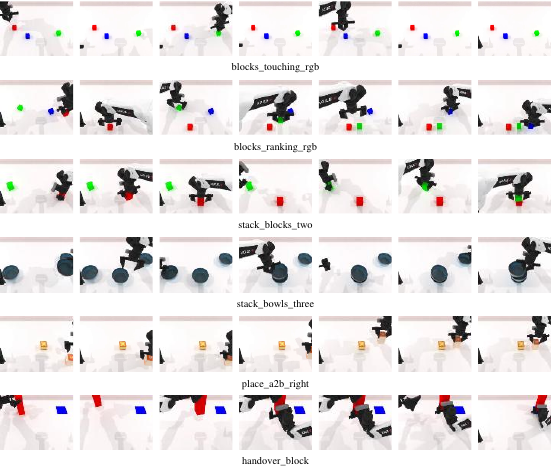}
    \vspace{-2mm}
    \caption{Demonstrations of the six tasks.}
    \label{fig:tasks}
    \vspace{-2mm}
\end{figure}

\clearpage
\section{Additional Qualitative Results}
\label{app:qualitative}

\begin{figure}[H]
  \centering
  \includegraphics[width=0.95\linewidth]{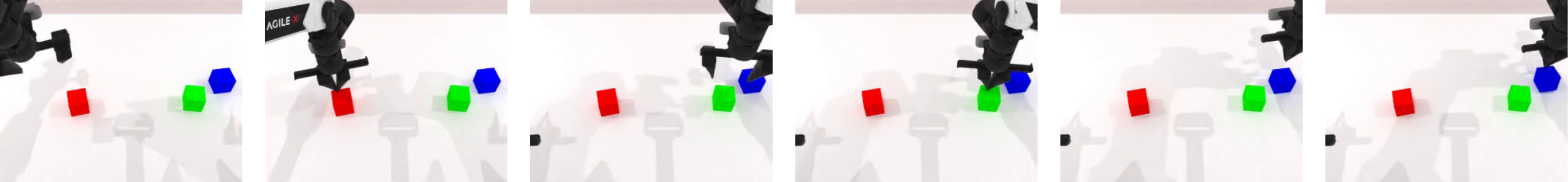}\\[.5mm]
  \includegraphics[width=0.95\linewidth]{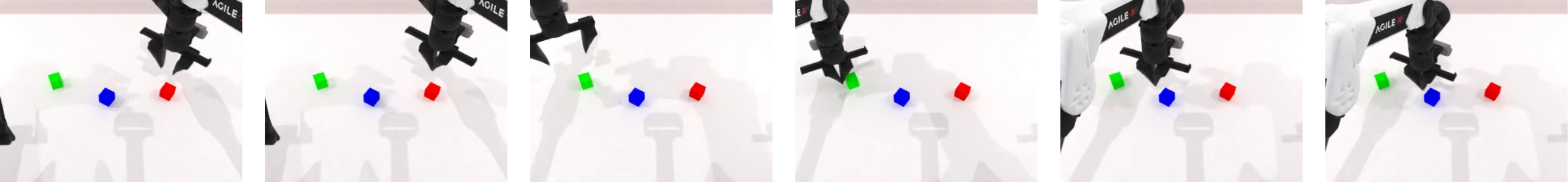}
  \vspace{-1mm}
  \caption{\textbf{Qualitative examples} on \textit{blocks\_touching\_rgb}.
  \textbf{Success (top):} the gripper touches the blocks in the required order.
  \textbf{Failure (bottom):} the gripper misses the red block and perturbs the green and blue blocks during contact.}
  \label{fig:blocks_touching_rgb_vis}
  \vspace{-2mm}
\end{figure}

\begin{figure}[H]
  \centering
  \includegraphics[width=\linewidth]{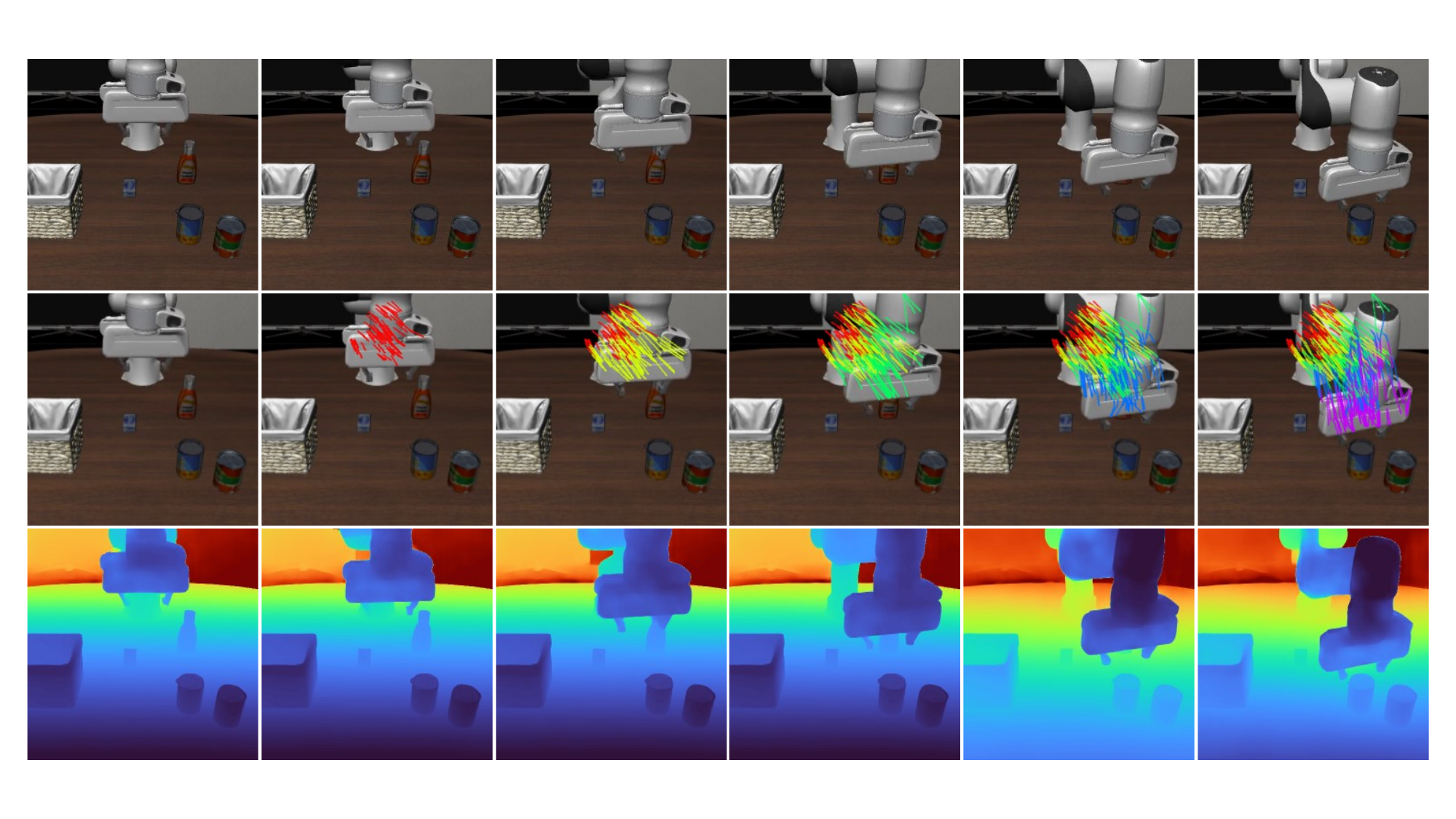}
  \vspace{-1mm}
  \caption{\textbf{LIBERO trajectory-field visualization.} Consecutive frames illustrate temporally structured motion cues extracted from the past observation window.}
  \label{fig:app_libero_traj_vis}
  \vspace{-2mm}
\end{figure}

\begin{figure}[H]
  \centering
  \includegraphics[width=\linewidth]{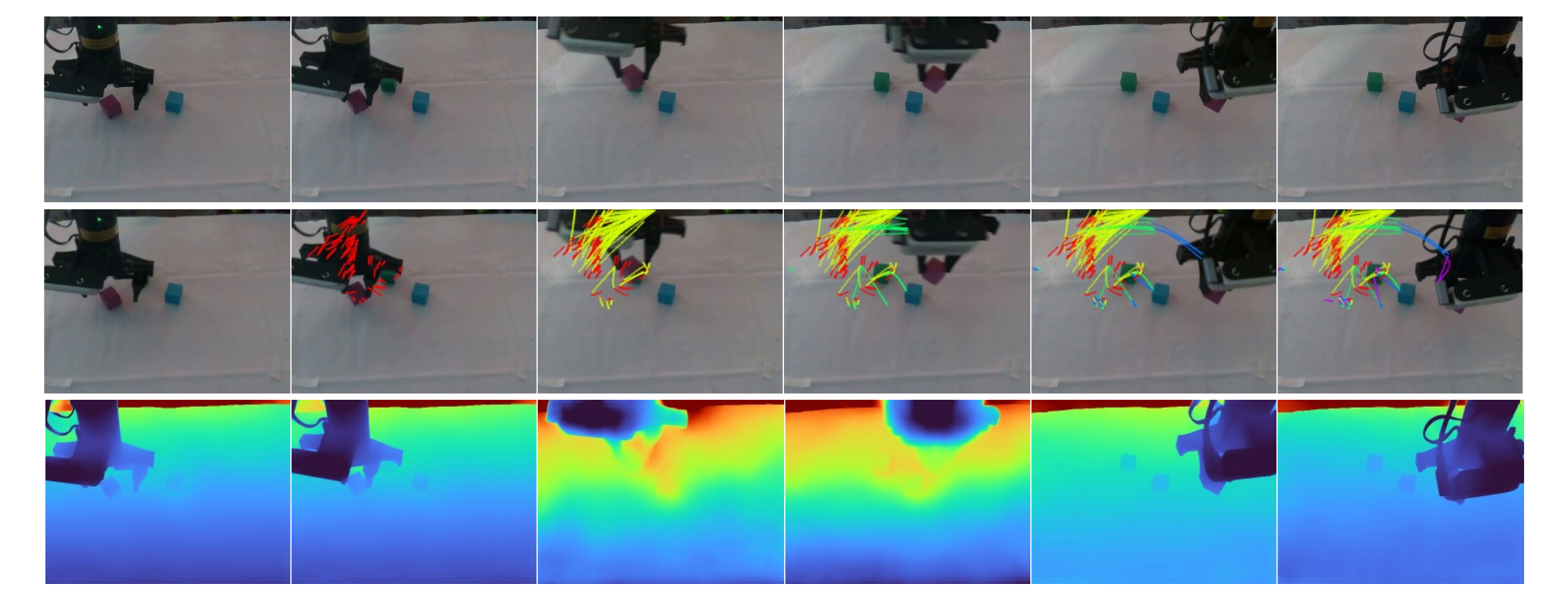}
  \vspace{-1mm}
  \caption{\textbf{Real-world trajectory-field visualization.} Consecutive Agilex Piper observations show the trajectory-field cues used by the motion-history interface.}
  \label{fig:app_real_traj_vis}
  \vspace{-2mm}
\end{figure}

\begin{figure}[H]
  \centering
  \includegraphics[width=0.95\linewidth]{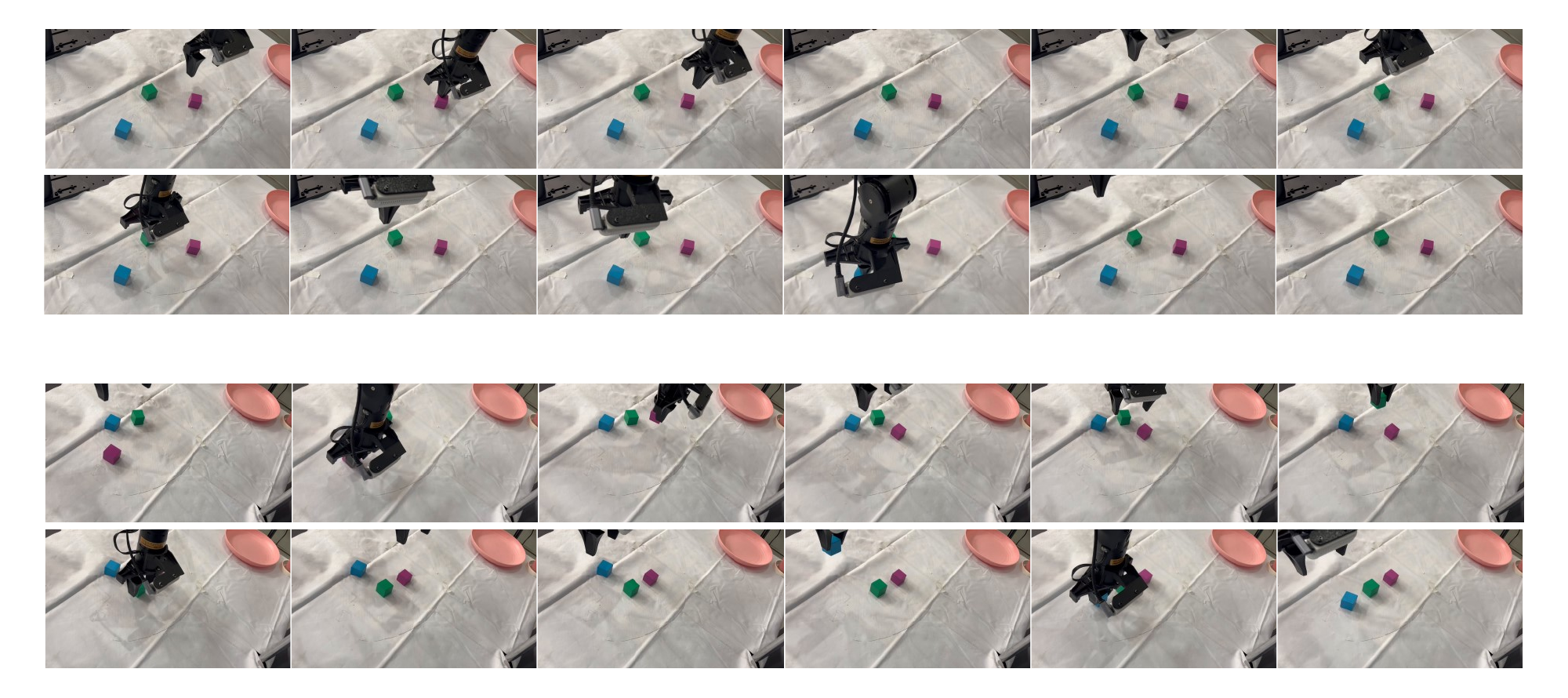}
  \vspace{-1mm}
  \caption{\textbf{Real-world rollouts.} Example rollouts on Agilex Piper for the preliminary real-world validation tasks.}
  \label{fig:app_real_rollouts}
\end{figure}

\end{document}